\documentclass[10pt,twocolumn,letterpaper]{article}

\usepackage{iccv}              

\usepackage[table]{xcolor}   
\usepackage{colortbl}        
\usepackage{booktabs}        
\usepackage{multirow}
\usepackage{graphicx}        
\definecolor{tablegray}{gray}{0.92} 
\usepackage{tcolorbox}
\usepackage{float}

\makeatletter
\renewcommand{\and}{\end{tabular}\hskip 1em \begin{tabular}[t]{c}}
\makeatother
\definecolor{iccvblue}{rgb}{0.21,0.49,0.74}
\usepackage[pagebackref,breaklinks,colorlinks,allcolors=iccvblue]{hyperref}

\def\paperID{*****} 
\def\confName{ICCV}
\def\confYear{2025}

\title{Bias-Aware Machine Unlearning: Towards Fairer Vision Models via Controllable Forgetting}

\author{
Sai Siddhartha Chary Aylapuram\\
BITS Pilani Dubai Campus\\
Dubai, UAE\\
{\tt\footnotesize h20230002@dubai.bits-pilani.ac.in }
\and
Veeraraju Elluru\\
IIT Jodhpur\\
India\\
{\tt\footnotesize b22cs080@iitj.ac.in }
\and
Shivang Agarwal\\
BITS Pilani Dubai Campus\\
Dubai, UAE\\
{\tt\footnotesize shivang@dubai.bits-pilani.ac.in }
}
\begin{document}
\maketitle
\begin{abstract}
Deep neural networks often rely on spurious correlations in training data, leading to biased or unfair predictions in safety-critical domains such as medicine and autonomous driving. While conventional bias mitigation typically requires retraining from scratch or redesigning data pipelines, recent advances in machine unlearning provide a promising alternative for post-hoc model correction. In this work, we investigate \textit{Bias-Aware Machine Unlearning}, a paradigm that selectively removes biased samples or feature representations to mitigate diverse forms of bias in vision models. Building on privacy-preserving unlearning techniques, we evaluate various strategies including Gradient Ascent, LoRA, and Teacher-Student distillation. Through empirical analysis on three benchmark datasets, CUB-200-2011 (pose bias), CIFAR-10 (synthetic patch bias), and CelebA (gender bias in smile detection), we demonstrate that post-hoc unlearning can substantially reduce subgroup disparities, with improvements in demographic parity of up to \textbf{94.86\%} on CUB-200, \textbf{30.28\%} on CIFAR-10, and \textbf{97.37\%} on CelebA. These gains are achieved with minimal accuracy loss and with methods scoring an average of 0.62 across the 3 settings on the joint evaluation of utility, fairness, quality, and privacy. Our findings establish machine unlearning as a practical framework for enhancing fairness in deployed vision systems without necessitating full retraining.
\end{abstract}
  
\section{Introduction}
\label{sec:intro}
\begin{figure}[ht]
\centering
\includegraphics[width=\linewidth]{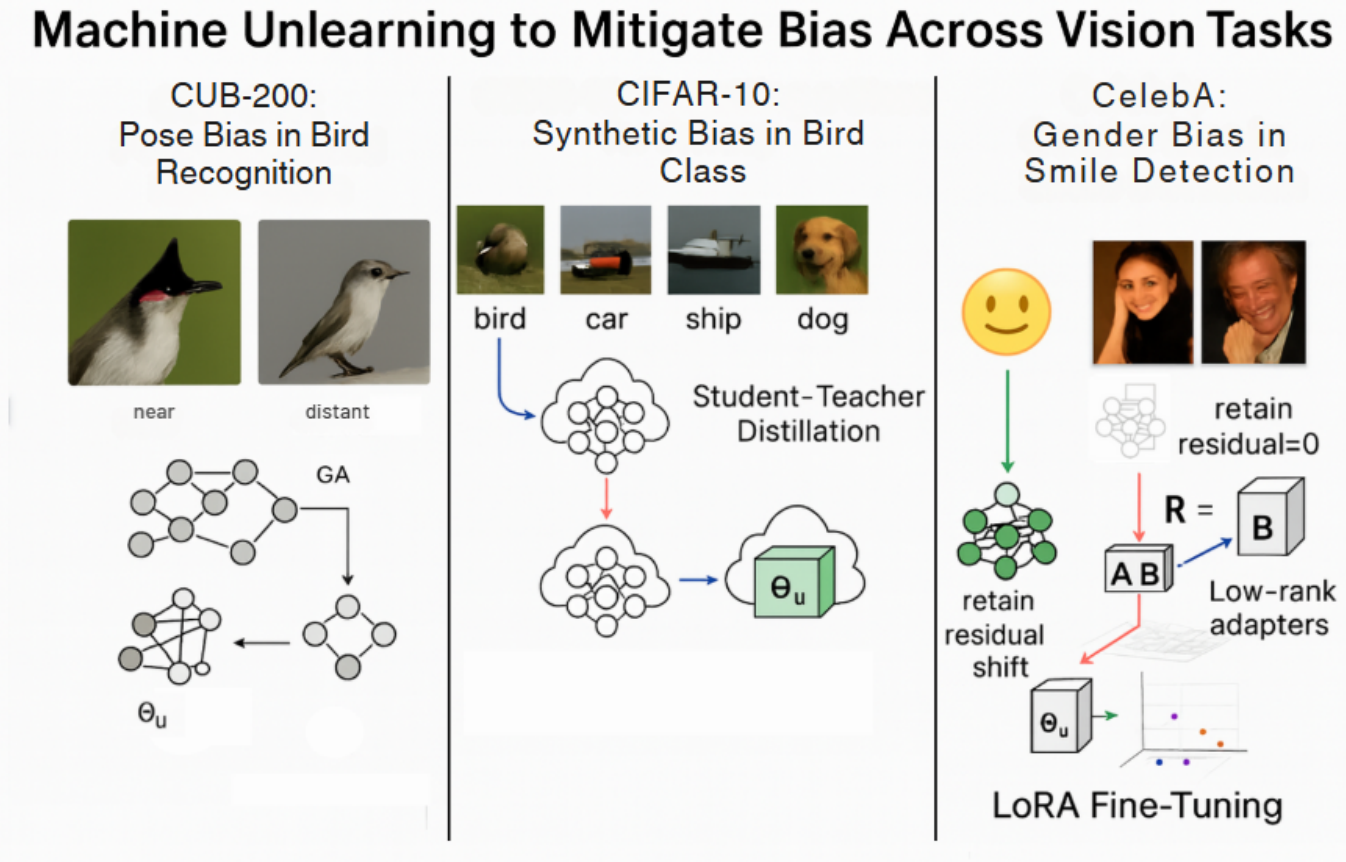}
\caption{%
\textbf{Machine Unlearning to Mitigate Bias Across Vision Tasks.}
This figure shows three representative debiasing examples.
(\textbf{Left}) CUB-200 (Pose bias in bird recognition): A gradient-ascent (GA) unlearning step debiases pose reliance by maximizing loss on spurious pose cues (e.g., near vs. distant) while preserving class-discriminative features.
(\textbf{Center}) CIFAR-10 (Forgetting a synthetic bias); The SCRUB framework, via student-teacher distillation, debiases the model with respect to a selected class (e.g., bird) by training the student only on retained classes and discarding information about the forgotten class.
(\textbf{Right}) CelebA (Gender bias in smile detection): LoRA fine-tuning debiases the gender-smile association using low-rank adapters; residual updates are constrained to low-rank matrices to unlearn biased correlations while maintaining task performance.
Together, these examples illustrate how unlearning can remove spurious correlations, protect privacy, and reduce demographic bias across vision models.
}
\label{fig:graphical_abstract}
\end{figure}
Deep learning models form the backbone of many high-performing computer vision systems used in real-world applications. However, numerous studies have shown that these models are prone to learning unintended biases due to spurious correlations in training data \cite{geirhos2020shortcut, torralba2011unbiased}. Such biases often manifest when a model over-relies on background textures, pose, or demographic attributes instead of learning semantically relevant features. These artifacts can severely impair model generalization and lead to discriminatory outcomes in real-world applications \cite{mehrabi2021survey}. 

To systematically investigate bias mitigation, it is essential to first establish a formal understanding of what constitutes \textit{bias} in visual recognition models and to delineate its various manifestations. We adopt the conventional supervised classification paradigm, in which the objective is to learn a mapping function \( f_\theta: \mathcal{X} \rightarrow \mathcal{Y} \), parameterized by \(\theta \in \Theta\), which maps an input image \( x \in \mathcal{X} \) to a corresponding class label \( y \in \mathcal{Y} \), following the foundational setup introduced by \citet{yenamandra2023factsamplifycorrelationsslice}. In real-world scenarios, the observed input \( x \) rarely provides a pure encoding of the task-relevant semantic signal \( s \); instead, it is more accurately modeled as a superposition:
\[
x = s + b
\]
Here, \( s \) denotes the intrinsic semantic information that is causally linked to the target label, whereas \( b \) represents spurious or biased features that are merely correlated with the label \( y \) in the training distribution. A model that relies disproportionately on \( b \), rather than extracting meaningful patterns from \( s \), is prone to failure under distributional shifts where the statistical association between \( b \) and \( y \) no longer holds \cite{dehdashtian2024fairnessbiasmitigationcomputer}. This phenomenon is commonly characterized as \textit{shortcut learning} \cite{geirhos2020shortcut}, where the model exploits superficial correlations instead of learning robust, generalizable representations, allowing the manifestation of bias.

\subsection{Bias in Vision Systems}
In modern vision systems, bias can manifest in several forms depending on the dataset and the overall learning setup, as illustrated in Figure \ref{fig:graphical_abstract}. Image classifiers have been observed to associate object categories with co-occurring background cues rather than object-specific features \cite{beery2018recognition, singh2020don}. For instance, in face recognition systems, biased training data have resulted in significantly lower accuracy for darker-skinned individuals and women \cite{buolamwini2018gender}. This behavior is especially problematic in high-stakes domains such as healthcare, or law enforcement, where fairness and robustness are critical. In this study, we analyze three representative types of bias:

\begin{itemize}
    \item \textbf{Pose Bias in CUB-200-2011}\footnote{\href{https://www.vision.caltech.edu/datasets/cub_200_2011/}{Caltech-UCSD Birds-200-2011 (CUB-200-2011)}}: Models trained on the CUB dataset tend to overfit on pose information (e.g., zoom level) rather than species-specific visual features. This occurs when certain poses are overrepresented in the training set, leading to a drop in accuracy in the underrepresented ones. \citet{branson2014birdspeciescategorizationusing} demonstrate that pose normalization schemes significantly affect classification accuracy, highlighting the risk of pose bias if diversity is not properly managed during training.
    
    \item \textbf{Class-Specific Bias in CIFAR-10}: Models may memorize certain classes more strongly than others, often due to imbalanced data augmentation or class distribution. This is especially problematic if a class must be removed for privacy or task-shift reasons, as the model may struggle to forget the targeted class without degrading performance on others. Such bias can be illustrated by introducing synthetic bias into a particular class to examine the effect of augmentation on the model’s predictions.
    
    \item \textbf{Attribute Bias in CelebA} \cite{liu2015faceattributes}: Demographic attributes such as gender can spuriously correlate with target labels (e.g., ‘Smiling’), leading to biased predictions. For instance, \citet{wang2020fairnessvisualrecognitioneffective} report an average gender skew of 80\% when the attribute ‘smiling’ is present.
\end{itemize}

\subsection{Machine Unlearning for Bias Mitigation}
Machine learning has been well established in the literature as being effective in mitigating a wide range of biases. This includes (i) \textbf{Sampling bias}, via downsampling overrepresented groups \cite{zhang2024geniurestricteddataaccess}, (ii) \textbf{Label bias}, by removing mislabeled or culturally skewed instances \cite{kodge2025sapcorrectivemachineunlearning}, (iii) \textbf{Proxy bias} through erasure of correlated but nonsensitive features \cite{xu2024dontforgetmuchmachine}, (iv) \textbf{Historical bias} by targeting systemic or source-specific biases \cite{DBLP:conf/emnlp/DigeAYZB0K24}, among many others. 

Current bias mitigation strategies can be broadly categorized as pre-processing (e.g., data augmentation), in-processing (e.g., adversarial training or reweighting), or post-processing techniques. Some of these techniques tend to require access to the entire training pipelines, may involve retraining from scratch, and often assume the bias structure is known a priori \cite{mehrabi2021survey, zhang2018mitigatingunwantedbiasesadversarial}. A recent work that employs a Fast Model Debiasing framework (FMD) \cite{chen2023fast} with machine unlearning, necessitates additional counterfactual dataset curation for debiasing. These limitations hinder their deployment in real-world settings largely comprising of pre-deployed models.

To address these challenges, we explore \emph{machine unlearning} (MU) as a promising post-hoc alternative for bias mitigation. Originally introduced in the context of data privacy and compliance with regulations such as GDPR \cite{bourtoule2021machine, ginart2019makingaiforgetyou}, MU seeks to remove the influence of specific training examples from a trained model without full retraining. We adapt this concept to fairness by selectively unlearning biased samples or spurious feature dependencies in trained models.

To formulate our problem setting, consider \(\mathcal{X}\), the input image space, and \(\mathcal{Y}\), the corresponding label space. Let \( \mathcal{D}_{\text{train}} = \{(x_i, y_i)\}_{i=1}^N \), where \( x_i \in \mathcal{X}, y_i \in \mathcal{Y}\), represent the training data distribution. A model trained on this data may learn a decision function \( f_\theta \) that overly relies on the bias component \( b \). In such cases, its predictions are predominantly influenced by \(b\) over the task-relevant, semantic signal \(s\), i.e.,

\[
\left\| \frac{\partial f_\theta(x)}{\partial b} \right\| \gg \left\| \frac{\partial f_\theta(x)}{\partial s} \right\|
\]

The objective of bias mitigation, therefore, is to reduce the model's reliance on \(b\) and to ensure that predictions are primarily driven by the task-relevant signal \(s\). As established in the literature \cite{oesterling2024fairmachineunlearningdata}, we define \textit{bias-aware unlearning} as the process of identifying a biased subset \( \mathcal{D}_b \subset \mathcal{D}_{\text{train}} \), and modifying the model such that its updated parameters \( \theta' \) approximate the behavior of a retrained model on the unbiased data:

\[
f_{\theta'} \approx f_{\theta^*}, \quad \text{where} \quad \theta^* = \arg\min_\theta \mathcal{L}(f_\theta, \mathcal{D}_{\text{train}} \setminus \mathcal{D}_b)
\]

This formulation enables us to study bias mitigation as a practical, post-hoc correction strategy that avoids retraining from scratch and aligns with real-world deployment needs. We study five popular techniques to achieve \textit{bias-aware machine unlearning} across various bias settings:
\begin{itemize}
    \item \textbf{Hard Unlearning}: Involves retraining the model from scratch on $\mathcal{D}_r$ to completely remove any influence of $\mathcal{D}_f$.
    \item \textbf{Gradient Ascent}: Utilizes a loss maximization strategy (via reversing the parameter update step) on $\mathcal{D}_f$ while regularizing to maintain performance on $\mathcal{D}_r$.
    \item \textbf{LoRA Fine-tuning}: This popular fine-tuning method introduces low-rank adaptation matrices to selectively forget $\mathcal{D}_f$ knowledge, in a parameter-efficient style.
    \item \textbf{Teacher-Student Unlearning}: \citet{kurmanji2023unboundedmachineunlearning} proposed a popular method that alternates between distillation away from the original model on \(\mathcal{D}_f\) and knowledge preservation using distillation towards the original model combined with a task-specific loss on \(\mathcal{D}_r\).
    \item \textbf{Counterfactual data-based Debiasing}: A strategy that involves identifying biased attributes using counterfactual examples and influence functions, which are removed by applying a lightweight unlearning update with only a small counterfactual dataset. Popularized by the work on Fast Model Debias with Machine Unlearning (FMD) \cite{chen2023fast}.
\end{itemize}

Lastly, building on concurrent multi-faceted evaluations such as MUSE \cite{shi2024muse} and GUM \cite{koudounas2025alexaforgetmemachine}, we explore bias-aware unlearning evaluation strategies that comprehensively evaluate MU methods on unlearning quality, post-unlearning bias, model efficiency and efficacy, via a single metric. 

We summarize our key contributions as follows:
\begin{itemize}
    \item We investigate the effectiveness of post-hoc unlearning strategies for mitigating bias in discriminative vision models, focusing on gradient ascent, LoRA fine-tuning, and teacher-student distillation-based methods, across a variety of bias types.
    
    \item We conduct a comparative study across three representative bias scenarios: pose bias in CUB-200, class forgetting in CIFAR-10, and attribute bias in CelebA, employing various unlearning strategies across each case.
    
    \item We benchmark these approaches using fairness-utility trade-offs (retain/forget accuracy, fairness gaps, concept activation vectors, and compute cost), and on a single unified metric, Concerted-Bias and Unlearning Metric (Co-BUM, read as ``Co-boom''), offering actionable insights for post-deployment model correction.
    
    \item We also qualitatively demonstrate the effectiveness of the chosen methods for bias mitigation via explainability-guided Grad-CAM visuals.
\end{itemize}

\section{Bias-Aware Machine Unlearning}
\label{sec:methods}

We describe the unlearning strategies employed across the three bias settings.

\noindent\underline{\textit{\textbf{Hard Unlearning}}}\label{methods:hard_unlearning}: Also called \emph{Exact Unlearning}, the goal of this method is to perform complete retraining on the retain dataset, \(\mathcal{D}_r\). This serves as a gold-standard oracle for model utility, post unlearning.

\noindent\underline{\textit{\textbf{Gradient Ascent}}}\label{methods:GA}: A model parameter-space intervention technique that removes specific learned associations by explicitly maximizing the loss on the forget set $\mathcal{D}_f$. Given a model $f_\theta$ with trained parameters $\theta_o$ trained on the full dataset $\mathcal{D}$, the goal is to adjust $\theta_o$ to $\theta_u$ such that $f_{\theta_u}$ performs poorly on $\mathcal{D}_f$ while retaining performance on $\mathcal{D}_r$. Following  \citet{DBLP:journals/corr/abs-2109-13398} the update step reads:
\[
\theta_{t+1} \leftarrow \theta_t + \eta \, \nabla_{\theta_t} (\mathcal{L}(\mathcal{D}_f; \theta_t) - \alpha.\mathcal{L}(\mathcal{D}_r; \theta_t)),
\]

where $\eta$ is the ascent learning rate, $\alpha$ is set to 1, and $\mathcal{L}(\cdot)$ denotes the task-specific loss, i.e., cross-entropy.

\noindent\underline{\textit{\textbf{LoRA Fine-Tuning}}}\label{methods:lora}: A popular parameter-efficient fine-tuning approach that isolates unlearning updates into Low-Rank Adaptation (LoRA) matrices, enabling selective removal of bias-associated knowledge while preserving most pre-trained weights. In LoRA, trainable matrices $(A, B)$ of rank $r \ll d$ are injected into existing weight matrices $W \in \mathbb{R}^{d \times k}$:
\[
W' = W + \Delta W,
\]
\[
where \quad \Delta W = A B, \quad A \in \mathbb{R}^{d \times r}, \, B \in \mathbb{R}^{r \times k}
\]
For unlearning, we freeze the original $W$ and optimize only $(A, B)$ to maximize the loss on $\mathcal{D}_f$ and minimize it on $\mathcal{D}_r$:
\[
\min_{A,B} \; \mathcal{L}(\mathcal{D}_r; W + AB) - \beta \, \mathcal{L}(\mathcal{D}_f; W + AB),
\]
where $\beta > 0$ scales the unlearning strength, set to 1 in our experiments. Once optimized, the LoRA parameters $(A, B)$ can be removed or replaced with alternative debiased adaptations, ensuring that the bias-associated behavior is no longer expressed in the model outputs while the majority of $W$ remains intact.

\noindent\underline{\textit{\textbf{Teacher-Student Unlearning}}}\label{methods:TSU}: This method frames the unlearning task in a teacher-student distillation paradigm. A model trained only on \(\mathcal{D}_r\) acts as the guiding, \emph{teacher} model with parameters $\theta_T$. Another model copy, called \textit{student}, with parameters $\theta_S$, is fine-tuned on \( \mathcal{D}_f\). For any input \( x \):
\[
\mathrm{let} \quad d(x; \theta_S) = D_{\mathrm{KL}}\bigl(p(f(x; \theta_T)) \,\big\|\, p(f(x; \theta_S))\bigr)
\]
be the KL divergence between the teacher’s softmax output \( p(f(x; \theta_T)) \) and the student’s \( p(f(x; \theta_S)) \). Drawing inspiration from the SCRUB framework, we initialize \( \theta_T \leftarrow \theta_S \) and then optimize a bi-objective during the fine-tuning process that:
\begin{itemize}
  \item 

\emph{Minimizes} \( d(x; \theta_S) \) for \( x \in \mathcal{D}_r \), encouraging the student to mimic the teacher on retained examples;
  \item 

\emph{Maximizes} \( d(x; \theta_S) \) for \( x \in \mathcal{D}_f \), forcing the student’s behavior to diverge from the teacher on forgotten examples.
\end{itemize}

This yields a single-phase training objective that balances the \textit{remembering} of general model utility and \textit{forgetting} of harmful bias. 

\noindent\underline{\textit{\textbf{Fast Model Debiasing (FMD)}}}\label{methods:FMD}: A post-hoc unlearning framework inspired from \citet{chen2023fast}, that identifies and removes bias from a trained model using a small counterfactual dataset, without necessitating original training data. The key steps involved are:
\begin{itemize}
    \item \textbf{Bias Quantification via Influence Functions.} Given a pretrained model $f_{\theta}$, and a bias, e.g., a spurious attribute correlating with predictions the influence of each sample $(x_i,y_i)$ on the bias measure is approximated using influence functions:
  \[
  \mathcal{I}(x_i) \approx - \nabla_\theta \ell(x_i,y_i)^\top H_\theta^{-1} \nabla_\theta B(\theta),
  \]
  where $\ell$ is the loss, $H_\theta$ is the Hessian of the empirical risk, and $B(\theta)$ quantifies model bias w.r.t the attribute.

  \item \textbf{Counterfactual Dataset.} A small external counterfactual dataset, $\mathcal{D}_c$, consists of samples where the spurious attribute is altered (e.g., bias removed or replaced), without access to the full original training set.

  \item \textbf{Unlearning Update.} A one-step Newton-style parameter update is computed to “unlearn” the biased influence of the content in $\mathcal{D}_c$:
  \[
  \theta_{t+1} = \theta_t - H_{\theta_t}^{-1} \left( \frac{1}{|\mathcal{D}_c|} \sum_{(x,y)\in \mathcal{D}_c} \nabla_{\theta_t} \ell(x,y) \right).
  \]
  This effectively reduces the estimated influence of the bias encoded in $f_{\theta}$ with minimal computational cost.
\end{itemize}

\section{Experimental Setup}
We evaluate the discussed unlearning-based bias mitigation strategies across three benchmark datasets, CUB-200-2011\footnote{\href{https://www.vision.caltech.edu/datasets/cub_200_2011/}{Caltech-UCSD Birds-200-2011 (CUB-200-2011)}}, CIFAR-10, and CelebA \cite{liu2015faceattributes}, each representing a distinct type of spurious correlation and bias. The evaluation is conducted using four complementary unlearning metrics. For each scenario, we describe the dataset preparation, model architecture, source of bias, and experimental setup in detail. For each task, we utilize the standard 70-10-20 dataset splits, corresponding to the train-val-test components respectively. We also employ the open-source version of the ResNet18-backbone model\footnote{\href{https://docs.pytorch.org/vision/main/models/generated/torchvision.models.resnet18.html}{PyTorch Models}} \cite{he2015deepresiduallearningimage}. The final, fully connected layer is adapted based on the dataset and overall experimental setting. For training purposes, the optimizer, learning rate, and the compute used were standardized, \texttt{Adam}, 1e-4, and an NVIDIA Tesla T4 GPU respectively.

\subsection{Pose Bias Mitigation on CUB-200-2011}

\textbf{Dataset.} The CUB-200-2011 dataset consists of $11,788$ bird images across $200$ species. The RGB images are resized to 224X224. The pose bias is not trivially observable. To investigate, we employed a normalized bounding box area as a proxy, partitioning it into three quantile-based bins representing different camera distances: Close-up, Mid-range, and Distant, denoted by discrete pose bins {0, 1, 2}. This revealed a pose-related bias - birds photographed from farther distances (or equivalently, containing small bounding boxes) often belong to certain classes, allowing shortcut learning. The bird samples with pose bins 0 and 1 constitute \(\mathcal{D}_r\), and those with pose bin 2, the \(\mathcal{D}_f\).

\noindent \textbf{Baseline.} We train a ResNet-18 for 10 epochs on the CUB training data split for task of bird class prediction. Suitable changes are made for the final linear layer for the 200-way classification.

\noindent \textbf{Bias Mitigation.} We evaluate five unlearning strategies:  \newline (i) Hard Unlearning, (ii) Gradient Ascent, (iii) LoRA-based Fine-Tuning: We utilize rank-8 LoRA adapters inserted at late convolutional  blocks (3-4), (iv) SCRUB: The (unbiased) teacher model is borrowed after the Exact Unlearning strategy, and the student model is initialized to the baseline, which is then fine-tuned according to the objective discussed in Section \ref{methods:TSU},  (v) FMD: The counterfactual dataset, \(\mathcal{D}_c\), is generated by resampling pose bins to ensure uniform pose distribution, and the classifier head is fine-tuned to suppress forget set predictions.

\subsection{Class-Specific Forgetting and Proxy Bias on CIFAR-10} 
\textbf{Dataset.} We introduce a synthetic shortcut bias in the CIFAR-10 dataset by overlaying a $8\times8$ red patch at the top-left corner of $50\%$ of the training samples from class-2 (`bird'). This creates a strong spurious correlation, where the presence of the patch can be exploited by the model instead of learning true class-discriminative features. This sets up \(\mathcal{D}_f\) as all the biased `bird + patch' samples and \(\mathcal{D}_r\) as the rest of the training set.

\noindent \textbf{Baseline.} The last layer of the ResNet-18, pretrained on ImageNet-1k, is fine-tuned on the CIFAR-10 training split (with the synthetic data) for 10 epochs by employing the \texttt{CrossEntropyLoss}.

\noindent \textbf{Bias Mitigation.} We benchmark five unlearning strategies against the (biased) baseline. All fine-tuning steps are done for 5 iterations unless stated otherwise. (i) Hard Unlearning: Retraining on \(\mathcal{D}_r\) for 10 epochs,  (ii) Gradient Ascent, (iii) LoRA-based Fine-tuning: We utilize low-rank adapters (Rank = 8) in the final convolutional block of ResNet-18 for fine-tuning on \(\mathcal{D}_r\), (iv) SCRUB: The (unbiased) teacher model directly results from the Exact Unlearning strategy, and the student model is initialized to the baseline, which is then fine-tuned according to the objective described in section \ref{methods:TSU}, (v) FMD: We construct counterfactual samples by masking the red patch. The baseline is then fine-tuned with a joint cross-entropy and contrastive embedding loss to enforce patch-invariance.

\subsection{Gender-``Smiling'' Attribute Correlation Unlearning in CelebA}

\textbf{Dataset.} CelebA consists of over $200,000$ celebrity face images annotated with attributes. The RGB images are resized to 128x128, post a \texttt{CenterCrop} augmentation. We perform binary classification on the `Smiling' attribute with `Gender' serving as the sensitive attribute (Male = 1, Female = 0). The training set is clearly imbalanced with respect to the identified gender-attribute correlation, i.e., there are more smiling females ($\sim$60000) than males ($\sim$10000), inducing a clear gender bias in smiling attribute predictions.

\noindent \textbf{Baseline Model.} We replace the final, fully connected layer of the backbone ResNet18 architecture to perform a 2-way classification on the sensitive attribute. We train this model by employing the \texttt{BinaryCrossEntropy} objective for 5 epochs, with a batch size of 64. Following previous experiments, this Vanilla ResNet18 model serves as the baseline.

\noindent \textbf{Bias Mitigation.} We evaluate four unlearning strategies on the CelebA dataset against the baseline ResNet-18. (i) Hard Unlearning: Retraining on \(\mathcal{D}_r\) for 10 epochs,  (ii) Gradient Ascent, (iii) LoRA-Fine-Tuning: Low-rank adapters (rank=4) are inserted into late convolutional blocks (3-4) for fine-tuning, (iv) SCRUB: A teacher trained only on the retain set provides soft targets; the student model is optimized with a distillation objective that aligns on retain data while diverging on forget data.
\newline\underline{\textit{Disclaimer:}} Due to the lack of trivially constructable counterfactuals, FMD is rendered irrelevant.

\begin{figure}[t]
\centering
\begin{tcolorbox}[colback=gray!10, colframe=black, boxrule=0.6pt,
  width=\columnwidth, arc=2pt, auto outer arc, top=4pt, bottom=4pt, left=6pt, right=6pt]

{\footnotesize
\textit{The Concerted Bias and Unlearning Metric (Co-BUM), combining unlearning utility ($U$), fairness ($F$), quality ($Q$), privacy ($P$), and efficiency ($E$) via a weighted harmonic mean. Normalization $\mathcal{N}_X$ scales deviations w.r.t. the gold model $M_g$ and baseline $M_0$, and penalizes regressions beyond baseline.}

\[
\begin{aligned}
& \text{Let } \mathcal{N}_X \text{ be the normalization score for a metric } X. \\
& \text{Then,} \\
& U = \tfrac{1}{2}\!\left[\tfrac{\mathrm{RA}(M_u)}{\mathrm{RA}(M_g)} 
      + \tfrac{\mathrm{TA}(M_u)}{\mathrm{TA}(M_g)}\right] \\
& F = 1 - \tfrac{1}{2}\left(\mathcal{N}_{\mathrm{DP}} 
      + \mathcal{N}_{\mathrm{EO}}\right), \quad P = 1 - \mathcal{N}_{\mathrm{MIA}}  \\
& Q = 1 - \tfrac{{\mathrm{FA}(M_u)}}{{\mathrm{FA}(M_g)}}, 
\quad E = \tfrac{\log T_g}{\log T_u} \\
\\
& \mathrm{Finally, Co\text{-}BUM} := \\
& \quad \kappa \left(\sum_{i \in \mathcal{S}} \alpha_i\right)
        \bigg/ \left(\sum_{i \in \mathcal{S}} \frac{\alpha_i}{i}\right),
        \text{ where } \mathcal{S} = \{U,F,P,Q,E\}
\end{aligned}
\]
}
\end{tcolorbox}
\caption{Definition of the Co-BUM metric.}
\label{fig:metric}
\end{figure}

\subsection{Evaluation Metrics}
Following earlier work on Controllable Machine Unlearning, ConMU \cite{xu2023controllable}, to assess the effectiveness of the aforementioned unlearning methods, we evaluate all strategies across the 3 datasets on:

\begin{itemize}
    \item \textbf{Unlearning Quality:} Measured as the classification accuracy on the forget set \(\mathcal{D}_f\), which comprises either the biased subgroup or the removed class, depending on the specific experimental scenario.
    
    \item \textbf{Model Utility:} Evaluated in terms of performance on the retained (and unbiased) subset \(\mathcal{D}_r\), post unlearning.

    \item \textbf{Unlearning Privacy:} Unlearning methods are prone to adversarial attacks \cite{DBLP:journals/corr/abs-2005-02205}. To measure this vulnerability, we report Membership Inference Attack (MIA) scores.

    \item \textbf{Model Fairness:} We report standard fairness metrics, including Equalized Odds \cite{hardt2016equality} and Demographic Parity \cite{Calders2009BuildingCW} to quantify disparities in model behavior across demographic partitions (e.g., male vs.\ female or far vs.\ near) to determine the model efficacy for debiasing.    
    \item \textbf{Unlearning Efficiency:} The total computational time for the post-hoc unlearning processes.
\end{itemize}

\begin{table}[H]
\centering
\footnotesize
\begin{tabular}{c c}
\toprule
Parameter & Value \\
\midrule
$\alpha_U$ & $0.25$ \\
$\alpha_F$ & $0.25$ \\
$\alpha_Q$ & $1$ \\
$\alpha_P$ & $1$ \\
$\alpha_E$ & $1$ \\
\midrule
$\gamma$   & $0.5$ \\
$\kappa$   & $1$ \\
\bottomrule
\end{tabular}
\caption{Weighting parameters ($\alpha$'s), asymmetry penalty $\gamma$, and the scaling factor $\kappa$ used in reporting Co-BUM scores.}
\label{tab:alphas}
\end{table}

\begin{table*}[h!]
\centering
\caption{Unlearning Results across CUB-200, CIFAR-10, and CelebA datasets. \newline 
$^{*}$FA: Forget Accuracy, 
$^{\dagger}$R.A.: Retain Accuracy, 
$^{\ddagger}$TA: Test Accuracy, 
$^{\rhd}$DP: Demographic Parity, 
$^{\diamond}$EO: Equalized Odds, 
$^{\spadesuit}$MIA: ROC-AUC of Membership Inference Attack. All fairness scores (DP and EO) are reported as \% drops w.r.t baseline model performance. Boldface indicates best performance and underline, the second-best. Runtime comparisons are only done amongst the unlearning methods and not w.r.t the Baseline}
\label{tab:unlearning_results}

\footnotesize
\resizebox{\textwidth}{!}{
\begin{tabular}{lcccccccc}
\toprule
\textbf{Method} & \textbf{FA$^{*}$ ($\downarrow$)} & \textbf{RA$^{\dagger}$ ($\uparrow$)} & \textbf{TA$^{\ddagger}$ ($\uparrow$)} & \textbf{DP$^{\rhd}$ (\%) ($\uparrow$)} & \textbf{EO$^{\diamond}$ (\%)($\uparrow$)} & \textbf{MIA$^{\spadesuit}$ ($\downarrow$)} & \textbf{Time (s)} ($\downarrow$) & \textbf{Co-BUM ($\uparrow$)}\\
\midrule

\multicolumn{9}{l}{\cellcolor{tablegray}\centering\textbf{CUB-200 (pose bias)}} \\[-1pt]
Baseline         & 80.67& 81.33& 78.69
&  0.00 & 0.00 & 0.56  & 519 & -- \\
\textcolor{gray}{Hard Unlearning}  & \textcolor{gray}{17.67} & \textcolor{gray}{72.33} & 63.03& \textcolor{gray}{94.51} & \textcolor{gray}{51.74} & \textcolor{gray}{0.48}  & \textcolor{gray}{222} & \textcolor{gray}{--} \\
Gradient Ascent  & \textbf{37}& 74.67& \underline{64.41}& \underline{93.75} & \underline{41.76} & \textbf{0.48}  & 299 & \textbf{0.71}\\
LoRA             & 77.83& \textbf{80.33}& 57.47& 88.53 & -11.14 & 0.52  & \textbf{233} &0.11\\
SCRUB            & 67.67& \underline{79.17}& 45.72& 91.92 & 8.53 & \underline{0.51}  & \underline{237} &0.38 \\
FMD              & \underline{66.83}& 69.83& \textbf{65.13}& \textbf{94.86} & \textbf{46.61} & 0.48  & 249 & \underline{0.52}\\

\midrule
\multicolumn{9}{l}{\cellcolor{tablegray}\centering\textbf{CIFAR-10 (synthetic bias)}} \\[-1pt]
Baseline         & 99.80 & 87.40 & 85.50 & 0.00 & 0.00 & 0.60 & 552 & --\\
\textcolor{gray}{Hard Unlearning} & \textcolor{gray}{14.10} & \textcolor{gray}{83.10} & \textcolor{gray}{75.70} & \textcolor{gray}{83.28} & \textcolor{gray}{83.28} & \textcolor{gray}{0.56}  & \textcolor{gray}{210} & \textcolor{gray}{--} \\
Gradient Ascent  & \textbf{45.30} & \underline{85.80} & 78.30 & -9.46 & -9.46 & 0.59  & \underline{266} &0.39 \\
LoRA             & \underline{48.70} & 84.80 & 79.00 & \textbf{30.28} & \textbf{30.28} & \textbf{0.58}  & \textbf{198} &\textbf{0.50} \\
SCRUB            & \textbf{45.30} & \textbf{86.30} & \textbf{84.70} & \underline{22.08} & \underline{22.08} & \underline{0.59}  & 323 &\underline{0.46}\\
FMD              & 74.70 & 85.80 & \underline{80.40} & -34.7 & -34.7 & 0.59  & 296 &0.38\\

\midrule
\multicolumn{9}{l}{\cellcolor{tablegray}\centering \textbf{CelebA (inter-attribute bias)}} \\[-1pt]
Baseline         & 95.95 & 96.13 & 94.20 & 0.00  & 0.00  & 0.67  & 1582 & -- \\
\textcolor{gray}{Hard Unlearning}  & \textcolor{gray}{87.27} & \textcolor{gray}{95.56} & \textcolor{gray}{92.33} & \textcolor{gray}{-29.62} & \textcolor{gray}{1.5} & \textcolor{gray}{0.51}  & \textcolor{gray}{1310} & \textcolor{gray}{--}\\
Gradient Ascent  &  \textbf{3.39} & 58.31 & 50.90 & \textbf{97.37}  & \textbf{98.13}  & \underline{0.51}  & \textbf{80} & \textbf{0.77} \\
    LoRA             & \underline{70.52} & \underline{87.81} & \underline{88.25} & -11.84  & \underline{18.77} & \textbf{0.47}  & 1024 & \underline{0.54} \\
SCRUB            & 79.24 & \textbf{93.89} & \textbf{91.01} & \underline{57.6}  & -71.05 & 0.53  & \underline{741} & 0.42 \\

\bottomrule
\end{tabular}
}
\end{table*}
\begin{figure*}[t]
  \centering
  \includegraphics[width=\textwidth]{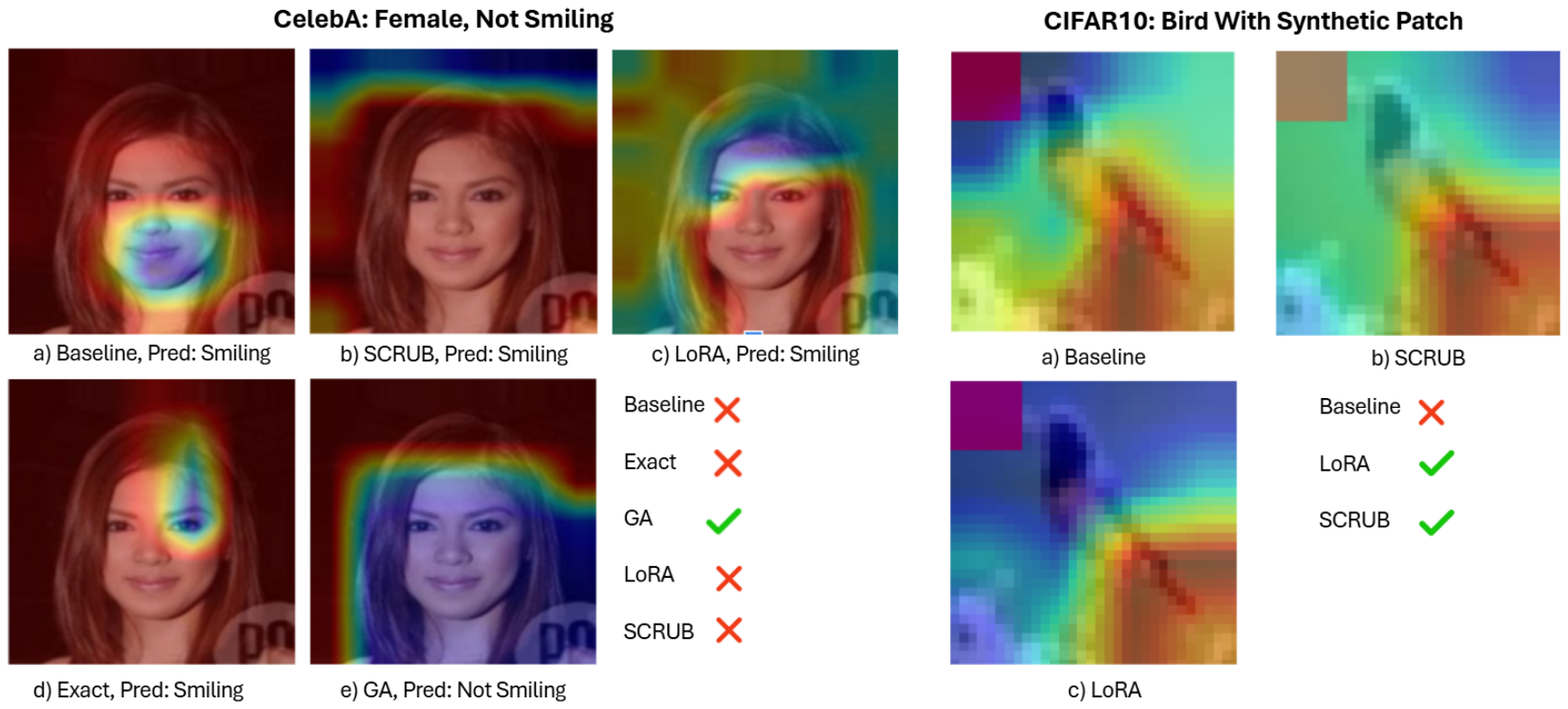}
  \caption{
    GradCAM activations showing the effectiveness of machine unlearning strategies. Bright purple/blue regions in Grad-CAM indicate higher contributions toward the model’s outputs.\newline \textbf{Left}: Samples containing the sensitive attribute pair (female-``smiling'') from the CelebA dataset (ID=055153).  Before Unlearning, the model heavily attends to biased facial regions - though the baseline has learned to understand cheek muscle cues, the overshadowing influence of gender is quite significant in the final prediction. Further, amongst the post-hoc unlearning methods b), c), and e), only GA is truly able to capture the facial structure and correctly classify as ``Not Smiling''. SCRUB does not pick up on any facial cues, but still predicts ``Smiling'' evidently due to the gender influence, as is with the Hard Unlearning strategy. \newline \textbf{Right}: Similarly, for the CIFAR-10 dataset (class 2), while the baseline method attends heavily to the red-patch, the LoRA and SCRUB methods show performant unlearning - they use features from the entirety of the image and not just the red patch.}
  \label{fig:gradcam_images}
\end{figure*}

Comprehensive evaluation of unlearning remains incomplete without accounting for these multifaceted objectives. As with other complex evaluation paradigms, no single scalar measure is sufficient to adjudicate the ``best'' model. Prior frameworks such as GUM \cite{koudounas2025alexaforgetmemachine}, MUSE \cite{shi2024muse}  highlight this gap by proposing composite scores that jointly capture efficacy, utility, and efficiency. Building on these efforts, we introduce a fairness-aware extension tailored to post-unlearning assessment. Specifically, our metric (Figure \ref{fig:metric}) systemically integrates bias-sensitive criteria (e.g., Equalized Odds, Demographic Parity) with unlearning-relevant factors (e.g., Forget Accuracy, Retain Accuracy, Membership Inference Attack resilience), all normalized against a gold-standard retrained model. This coupling provides a principled, unified index that quantifies not only how well a model selectively forgets, but also how equitably it behaves once unlearning has been performed. The co-efficients are summarized in Table \ref{tab:alphas}.

\section{Results}
We evaluate our unlearning approaches on three benchmark vision datasets (Table \ref{tab:unlearning_results}), CUB-200, CIFAR-10, and CelebA, each representing different forms of bias.

\noindent \underline{\textit{General Trends:}} Our evaluation shows the need to jointly consider unlearning and fairness in bias mitigation. Unlearning removes spurious correlations, while fairness metrics reveal which correlations are excised and whether subgroup equity is achieved. Across datasets, we consistently observe divergence between demographic parity (DP) and equalized odds (EO): DP often improves markedly, redistributing positive predictions across groups, yet EO lags, indicating persistent error-rate imbalances despite seemingly fair prediction rates. This distinction shows fairness gains may be superficial rather than substantive. The RA-TA gap further serves as a diagnostic, small gaps signal targeted forgetting with minimal collateral damage, while large gaps indicate indiscriminate degradation. Co-BUM trends help adjudicate methods that balance across competing objectives.

\noindent \underline{\textit{Mitigating Pose bias:}} Pose bias is subtle and distributed across representations rather than localized. Methods that explicitly push the decision boundary against the forget set (like GA) achieve decisive forgetting and large DP gains but only moderate EO improvements, showing that removing shortcuts does not fully equalize group error rates. Counterfactual rebalancing provides a steadier trade-off, achieving forgetting with a narrow RA-TA gap, evidencing selective rather than indiscriminate adjustment. On the flip side, parameter-efficient fine-tuning struggles since the late-layer adapters fail to disentangle class predictions from pose, (evidenced by a high RA, but poor TA and EO). Thus, for diffuse biases like pose, boundary-pushing methods are preferable. These trends are captured by Co-BUM, rewarding their alignment of forgetting, fairness, utility, and privacy.
\newline \newline
\noindent \underline{\textit{Mitigating Synthetic bias:}} The synthetic red-patch bias is sharply localized and easily exploited. In this setting (contrary to the previous), LoRA is highly effective because small targeted updates suppress reliance on the patch while preserving class-discriminative capacity with minimal memorization (lowest MIA). This yields strong fairness gains and competitive generalization. Teacher-student distillation also approaches baseline accuracy but undercorrects the bias, questioning the fit-for-use from the fairness perspective. Gradient-based forgetting is less effective because without the explicit constraints, the model finds auxiliary cues tied to the patch, allowing the shortcut to persist. Asymmetric counterfactual fine-tuning can even backfire, since the absence of the patch becomes predictive of class membership (another direct shortcut). Thus, for localized biases, parameter-efficient tuning best balances bias removal and task quality. Counterfactual strategies must be symmetric to avoid new shortcuts. Co-BUM highlights that small, localized updates suppress the synthetic bias while preserving generalization and is further justified in the GradCAM images in Figure \ref{fig:gradcam_images} (Right) that show the part of the image that contributes towards the model's predictions.
\newline \newline
\noindent \underline{\textit{Mitigating Inter-attribute bias:}} The gender-smiling correlation reflects a socially sensitive, entrenched bias. Aggressive unlearning strategies sharply reduce disparities, nearly eliminating both DP and EO gaps, but at the cost of steep performance decline, highlighting over-forgetting. LoRA-based fine-tuning offers a pragmatic compromise: it mitigates the correlation, improves EO, and preserves test accuracy near baseline, yet leaves residual disparities, particularly in DP, underscoring its limits against entrenched biases. Distillation reveals a subtler risk - it can equalize prediction rates (DP) while retaining asymmetric errors (EO), since the retain-only teacher transfers its imbalanced error distribution to the student. Thus, it should only be applied with fairness-aware teachers explicitly corrected for EO disparities. Co-BUM ranks gradient ascent highest, reflecting that, with entrenched inter-attribute correlations, it prioritizes fairness removal and subgroup stability even at the expense of utility (Figure \ref{fig:gradcam_images}).

\section{Conclusion} We evaluated unlearning-based bias mitigation across CUB-200, CIFAR-10, and CelebA, encompassing distributed, localized, and socially entrenched biases. Boundary-focused methods and counterfactual rebalancing proved most effective for distributed correlations such as pose, parameter-efficient fine-tuning excelled for localized artifacts, and entrenched inter-attribute biases demanded an explicit trade-off between fairness and utility. Privacy improvements generally accompanied fairness gains, showing alignment. To adjudicate these competing objectives, we also evaluated on a composite metric (Co-BUM) that rewards balanced improvements across forgetting, fairness, utility, and privacy. These findings emphasize that bias mitigation is inherently context-sensitive and that systematic evaluation with Co-BUM provides actionable guidance for selecting appropriate unlearning strategies.

\textbf{Future Work.} We motivate readers to focus on developing adaptive unlearning methods that adjust to the topology of bias, whether distributed, localized, or entrenched, and that explicitly integrate fairness and privacy considerations. Promising directions include leveraging concept salience, mechanistic interpretability, and group disparity diagnostics to guide unlearning, as well as extending to multi-bias disentanglement. Equal emphasis should be placed on designing scalable, fairness-aware training pipelines to ensure reliability, transparency, and applicability in real-world, privacy-sensitive settings.
{
    \small
    \bibliographystyle{ieeenat_fullname}
    \bibliography{main}

\begin{thebibliography}{29}
\providecommand{\natexlab}[1]{#1}
\providecommand{\url}[1]{\texttt{#1}}
\expandafter\ifx\csname urlstyle\endcsname\relax
  \providecommand{\doi}[1]{doi: #1}\else
  \providecommand{\doi}{doi: \begingroup \urlstyle{rm}\Url}\fi

\bibitem[Beery et~al.(2018)Beery, Van~Horn, and Perona]{beery2018recognition}
Sara Beery, Grant Van~Horn, and Pietro Perona.
\newblock Recognition in terra incognita.
\newblock In \emph{(ECCV)}, 2018.

\bibitem[Bourtoule et~al.(2021)Bourtoule, Chandrasekaran, et~al.]{bourtoule2021machine}
Ludovic Bourtoule, Varun Chandrasekaran, et~al.
\newblock Machine unlearning.
\newblock In \emph{IEEE Symposium on Security and Privacy}, 2021.

\bibitem[Branson et~al.(2014)Branson, Van~Horn, Belongie, and Perona]{branson2014birdspeciescategorizationusing}
Steve Branson, Grant Van~Horn, Serge Belongie, and Pietro Perona.
\newblock Bird species categorization using pose normalized deep convolutional nets, 2014.

\bibitem[Buolamwini and Gebru(2018)]{buolamwini2018gender}
Joy Buolamwini and Timnit Gebru.
\newblock Gender shades: Intersectional accuracy disparities in commercial gender classification.
\newblock In \emph{Conference on fairness, accountability and transparency}. PMLR, 2018.

\bibitem[Calders et~al.(2009)Calders, Kamiran, and Pechenizkiy]{Calders2009BuildingCW}
Toon Calders, Faisal Kamiran, and Mykola Pechenizkiy.
\newblock Building classifiers with independency constraints.
\newblock In \emph{{ICDM} Workshops}, 2009.

\bibitem[Chen et~al.(2021)Chen, Zhang, Wang, Backes, Humbert, and Zhang]{DBLP:journals/corr/abs-2005-02205}
Min Chen, Zhikun Zhang, Tianhao Wang, Michael Backes, Mathias Humbert, and Yang Zhang.
\newblock When machine unlearning jeopardizes privacy.
\newblock In \emph{{ACM} {SIGSAC}}, 2021.

\bibitem[Chen et~al.(2023)Chen, Yang, Xiong, Bai, Hu, Hao, Feng, Zhou, Wu, and Liu]{chen2023fast}
Ruizhe Chen, Jianfei Yang, Huimin Xiong, Jianhong Bai, Tianxiang Hu, Jin Hao, Yang Feng, Joey~Tianyi Zhou, Jian Wu, and Zuozhu Liu.
\newblock Fast model debias with machine unlearning.
\newblock In \emph{NeurIPS}, 2023.

\bibitem[Dehdashtian et~al.(2024)Dehdashtian, He, Li, Balakrishnan, Vasconcelos, Ordonez, and Boddeti]{dehdashtian2024fairnessbiasmitigationcomputer}
Sepehr Dehdashtian, Ruozhen He, Yi Li, Guha Balakrishnan, Nuno Vasconcelos, Vicente Ordonez, and Vishnu~Naresh Boddeti.
\newblock Fairness and bias mitigation in computer vision: A survey, 2024.

\bibitem[Dige et~al.(2024)Dige, Arneja, Yau, Zhang, Bolandraftar, Zhu, and Khattak]{DBLP:conf/emnlp/DigeAYZB0K24}
Omkar Dige, Diljot Arneja, Tsz~Fung Yau, Qixuan Zhang, Mohammad Bolandraftar, Xiaodan Zhu, and Faiza~Khan Khattak.
\newblock Can machine unlearning reduce social bias in language models?
\newblock In \emph{{EMNLP}}, pages 954--969, 2024.

\bibitem[Geirhos et~al.(2020)Geirhos, Jacobsen, Michaelis, Zemel, Brendel, Bethge, and Wichmann]{geirhos2020shortcut}
Robert Geirhos, J{\"o}rn-Henrik Jacobsen, Claudio Michaelis, Richard Zemel, Wieland Brendel, Matthias Bethge, and Felix~A Wichmann.
\newblock Shortcut learning in deep neural networks.
\newblock \emph{Nature Machine Intelligence}, 2\penalty0 (11):\penalty0 665--673, 2020.

\bibitem[Ginart et~al.(2019)Ginart, Guan, Valiant, and Zou]{ginart2019makingaiforgetyou}
Antonio Ginart, Melody Guan, Gregory Valiant, and James~Y Zou.
\newblock Making ai forget you: Data deletion in machine learning.
\newblock \emph{NeurIPS}, 2019.

\bibitem[Hardt et~al.(2016)Hardt, Price, and Srebro]{hardt2016equality}
Moritz Hardt, Eric Price, and Nathan Srebro.
\newblock Equality of opportunity in supervised learning.
\newblock In \emph{(NeurIPS)}, 2016.

\bibitem[He et~al.(2015)He, Zhang, Ren, and Sun]{he2015deepresiduallearningimage}
Kaiming He, Xiangyu Zhang, Shaoqing Ren, and Jian Sun.
\newblock Deep residual learning for image recognition, 2015.

\bibitem[Kodge et~al.(2025)Kodge, Ravikumar, Saha, and Roy]{kodge2025sapcorrectivemachineunlearning}
Sangamesh Kodge, Deepak Ravikumar, Gobinda Saha, and Kaushik Roy.
\newblock {SAP:} corrective machine unlearning with scaled activation projection for label noise robustness.
\newblock In \emph{AAAI}, 2025.

\bibitem[Koudounas et~al.(2025)Koudounas, Savelli, Giobergia, and Baralis]{koudounas2025alexaforgetmemachine}
Alkis Koudounas, Claudio Savelli, Flavio Giobergia, and Elena Baralis.
\newblock "alexa, can you forget me?" machine unlearning benchmark in spoken language understanding, 2025.

\bibitem[Kurmanji et~al.(2023)Kurmanji, Triantafillou, Hayes, and Triantafillou]{kurmanji2023unboundedmachineunlearning}
Meghdad Kurmanji, Peter Triantafillou, Jamie Hayes, and Eleni Triantafillou.
\newblock Towards unbounded machine unlearning.
\newblock In \emph{NeurIPS}, 2023.

\bibitem[Liu et~al.(2015)Liu, Luo, Wang, and Tang]{liu2015faceattributes}
Ziwei Liu, Ping Luo, Xiaogang Wang, and Xiaoou Tang.
\newblock Deep learning face attributes in the wild.
\newblock In \emph{(ICCV)}, 2015.

\bibitem[Liu et~al.(2024)Liu, Dou, Chien, Zhang, Tian, and Zhu]{xu2023controllable}
Zheyuan Liu, Guangyao Dou, Eli Chien, Chunhui Zhang, Yijun Tian, and Ziwei Zhu.
\newblock Breaking the trilemma of privacy, utility, and efficiency via controllable machine unlearning.
\newblock In \emph{{WWW}}. {ACM}, 2024.

\bibitem[Mehrabi et~al.(2021)Mehrabi, Morstatter, Saxena, Lerman, and Galstyan]{mehrabi2021survey}
Ninareh Mehrabi, Fred Morstatter, Nripsuta Saxena, Kristina Lerman, and Aram Galstyan.
\newblock A survey on bias and fairness in machine learning.
\newblock \emph{ACM computing surveys (CSUR)}, 54\penalty0 (6):\penalty0 1--35, 2021.

\bibitem[Oesterling et~al.(2024)Oesterling, Ma, Calmon, and Lakkaraju]{oesterling2024fairmachineunlearningdata}
Alex Oesterling, Jiaqi Ma, Flavio Calmon, and Himabindu Lakkaraju.
\newblock Fair machine unlearning: Data removal while mitigating disparities.
\newblock In \emph{International Conference on Artificial Intelligence and Statistics}. PMLR, 2024.

\bibitem[Shi et~al.(2025)Shi, Lee, Huang, Malladi, Zhao, Holtzman, Liu, Zettlemoyer, Smith, and Zhang]{shi2024muse}
Weijia Shi, Jaechan Lee, Yangsibo Huang, Sadhika Malladi, Jieyu Zhao, Ari Holtzman, Daogao Liu, Luke Zettlemoyer, Noah~A. Smith, and Chiyuan Zhang.
\newblock {MUSE:} machine unlearning six-way evaluation for language models.
\newblock In \emph{{ICLR}}, 2025.

\bibitem[Singh et~al.(2020)Singh, Mahajan, Grauman, Lee, Feiszli, and Ghadiyaram]{singh2020don}
Krishna~Kumar Singh, Dhruv Mahajan, Kristen Grauman, Yong~Jae Lee, Matt Feiszli, and Deepti Ghadiyaram.
\newblock Don't judge an object by its context: learning to overcome contextual bias.
\newblock In \emph{CVPR}, 2020.

\bibitem[Thudi et~al.(2021)Thudi, Deza, Chandrasekaran, and Papernot]{DBLP:journals/corr/abs-2109-13398}
Anvith Thudi, Gabriel Deza, Varun Chandrasekaran, and Nicolas Papernot.
\newblock Unrolling {SGD:} understanding factors influencing machine unlearning.
\newblock \emph{CoRR}, abs/2109.13398, 2021.

\bibitem[Torralba and Efros(2011)]{torralba2011unbiased}
Antonio Torralba and Alexei~A Efros.
\newblock Unbiased look at dataset bias.
\newblock In \emph{CVPR}, 2011.

\bibitem[Wang et~al.(2020)Wang, Qinami, Karakozis, Genova, Nair, Hata, and Russakovsky]{wang2020fairnessvisualrecognitioneffective}
Zeyu Wang, Klint Qinami, Ioannis~Christos Karakozis, Kyle Genova, Prem Nair, Kenji Hata, and Olga Russakovsky.
\newblock Towards fairness in visual recognition: Effective strategies for bias mitigation.
\newblock In \emph{{CVPR}}, 2020.

\bibitem[Xu et~al.(2025)Xu, Zhu, Zhou, and Zhao]{xu2024dontforgetmuchmachine}
Heng Xu, Tianqing Zhu, Wanlei Zhou, and Wei Zhao.
\newblock Don't forget too much: Towards machine unlearning on feature level.
\newblock \emph{{IEEE} Trans. Dependable Secur. Comput.}, 22\penalty0 (2):\penalty0 1313--1328, 2025.

\bibitem[Yenamandra et~al.(2023)Yenamandra, Ramesh, Prabhu, and Hoffman]{yenamandra2023factsamplifycorrelationsslice}
Sriram Yenamandra, Pratik Ramesh, Viraj Prabhu, and Judy Hoffman.
\newblock Facts: First amplify correlations and then slice to discover bias, 2023.

\bibitem[Zhang et~al.(2018)Zhang, Lemoine, and Mitchell]{zhang2018mitigatingunwantedbiasesadversarial}
Brian~Hu Zhang, Blake Lemoine, and Margaret Mitchell.
\newblock Mitigating unwanted biases with adversarial learning.
\newblock In \emph{{AIES}}. {ACM}, 2018.

\bibitem[Zhang et~al.(2024)Zhang, Shen, Zhao, Chen, and Xu]{zhang2024geniurestricteddataaccess}
Chenhao Zhang, Shaofei Shen, Yawen Zhao, Weitong~Tony Chen, and Miao Xu.
\newblock Geniu: A restricted data access unlearning for imbalanced data, 2024.

\end{thebibliography}
}

\end{document}